\documentclass[letterpaper, 10 pt, conference]{ieeeconf}  

\IEEEoverridecommandlockouts                              

\usepackage{graphics} 
\usepackage{epsfig} 
\usepackage{mathptmx} 
\usepackage{times} 
\usepackage{amsmath} 
\usepackage{amssymb}  
\usepackage{booktabs}
\usepackage{siunitx}
\usepackage{multirow}
\usepackage{xcolor}
\usepackage[noadjust]{cite}
\newcommand{\updatedtext}[1]{\textcolor{black}{#1}}

\title{\LARGE \bf
Toward Force Estimation in Robot-Assisted Surgery\\ using Deep Learning with Vision and Robot State 
}

\author{Zonghe Chua$^{1}$, Anthony M. Jarc$^{2}$, and Allison M. Okamura$^{1}$
\thanks{$^{1}$Z. Chua and A. M. Okamura are with Department of Mechanical Engineering, Stanford University, Stanford, CA, 94305. \qquad {\tt\small chuazh@stanford.edu}, {\tt\small aokamura@stanford.edu}
        }%
\thanks{$^{2}$A. M. Jarc is with Intuitive Surgical Inc., Sunnyvale, CA, 94086. {\tt\small anthony.jarc@intusurg.com}
        }%
\thanks{*This work was supported in part by a Stanford Bio-X and Stanford Interdisciplinary Graduate Fellowship.}
}

\begin{document}

\maketitle
\thispagestyle{empty}
\pagestyle{empty}

\begin{abstract}

Knowledge of interaction forces during teleoperated robot-assisted surgery could be used to enable force feedback to users and evaluate tissue handling skill. However, direct force sensing at the end-effector is challenging \updatedtext{because} it requires biocompatible, sterilizable, and cost-effective sensors. Vision-based \updatedtext{neural networks} are a promising approach for providing useful force estimates, though questions remain about generalization to new scenarios and real-time inference. We present a force estimation neural network that uses RGB images and robot state as inputs. Using a self-collected dataset, we compared the network to variants that included only a single input type, and evaluated how they generalized to new viewpoints, workspace positions, materials, and tools. \updatedtext{We found that the vision-only network was sensitive to shifts in viewpoints, while networks with state inputs were sensitive to vertical shifts in workspace. The network with both state and vision inputs had the highest accuracy for an unseen tool, while the state-only network was most accurate for an unseen material.} Through feature removal studies, we found that using only \updatedtext{force features produced better accuracy than using only kinematic features as input.} The network with both state and vision inputs outperformed a physics-based model in accuracy for seen material. It showed comparable accuracy but faster computation times than a recurrent neural network, making it better suited for real-time applications.

\end{abstract}

\section{INTRODUCTION}

Safe and effective tissue handling is an important skill in robot-assisted minimally invasive surgery (RMIS) because it allows a surgeon to manipulate tissue without causing trauma or excess bleeding. One approach to improving tissue handling and force estimation skill is to provide real-time haptic feedback to the surgeon. This has been shown to reduce excessive interaction forces in dry lab tasks \cite{Wagner2007,Talasaz2017,Demi2005}. In such situations, feedback can be provided by force sensors placed in the environment \cite{Talasaz2017} or at the end-effector \cite{Wagner2007}\cite{Demi2005}\cite{Kuebler2005}\cite{Kim2015}, with the latter having potential for in vivo use. However, such technology has yet to see commercial adoption because it is challenging for force sensors to meet the requirements of biocompatibility, sterilizability, and cost-effectiveness \cite{Enayati2016}.

An alternative approach to improve surgeons' tissue handling and force modulation abilities is through skill assessment and feedback. Traditionally, this has been done via expert video review using qualitative rubrics \cite{goh2012global}. This method suffers from issues in rating consistency, timeliness of feedback, and time demands on raters.
Another approach to evaluation has been automated performance metrics (APMs) that are quantitative measures of skill \cite{Richards2000apms,Brown2017apms,hung2018utilizing,Hung2018apms2}. While APMs for skill categories such as economy of motion and bimanual dexterity are well known, and easily measurable by querying the robot state and kinematics, APMs for measuring tissue handling are difficult to implement without either environmental or end-effector force sensing, leading to force APMs being excluded from clinical studies \cite{Hung2018apms2}. 

There have been several attempts to estimate forces while circumventing the need for distal force sensing. A traditional approach to force estimation at the end-effector has been to use physics-based models and estimated joint torques \cite{fontanelli2017modelling,Wang2019dynamic,Pique2019dynamic}. However, due to the dynamics of the cable-driven tools, these models do not generalize well outside the set of trajectories used to fit their parameters. More recently, neural networks have been shown to be an alternative to physics-based models for estimating end-effector forces, achieving comparable results \cite{Yilmaz2020dynamic}. 

Besides robot state inputs, vision can be useful for force estimation. 
This is akin to how surgeons might estimate forces both while performing and assessing, surgery. Despite the lack of haptic feedback, they may use a prior understanding of environment stiffness, with visually and physically perceived displacement, to perform force estimation. 
In one approach, stereo images were used to construct a depth map of the environment, which was mapped to a finite element mesh with properties consistent with the manipulated tissue \cite{Haouchine2018FEA}. In an ex vivo experiment, this method correctly approximated the trends of the interaction forces, but with an offset and scaling \cite{Haouchine2018FEA}. 

Image-based convolutional neural networks trade off the requirement of explicit model specification needed in finite element approaches, with the need for more labeled data. Convolutional neural networks for vision-based force estimation have been explored as recurrent \cite{aviles2014recurrent,aviles2016towards,Lee2018lstm,Marban2018,Marban2019rnn,Jung2021,Lee2020Sages} or attentional networks \cite{Shin2019transformer} that can encode temporal information. Of these networks, some leverage robot state information, such as joint currents \cite{Lee2018lstm} and end-effector position \cite{aviles2016towards}\cite{Marban2019rnn}\cite{Jung2021}. The data-driven nature of these networks introduces concerns about generalizability to unseen scenarios that could be encountered in surgical practice. These can include unseen viewpoints, robot configurations, tools and materials. \updatedtext{While approaches that estimate the deformation of tissue using stereo lattice reconstruction can address limitations in tool and viewpoint generalization \cite{aviles2014recurrent}\cite{aviles2016towards}, they involve additional stereo image processing steps, computationally intensive optimization, and sequential input processing that} raise concerns about their suitability for providing real-time feedback, even for the relatively slow speeds required to implement visual indicators. \updatedtext{Such image processing and sequential computation is also found in monocular approaches that attempt to incorporate temporal history \cite{Marban2019rnn}.}



\begin{figure}[!t]
\vspace{1.5mm}
\centering
\includegraphics[width=0.7\linewidth]{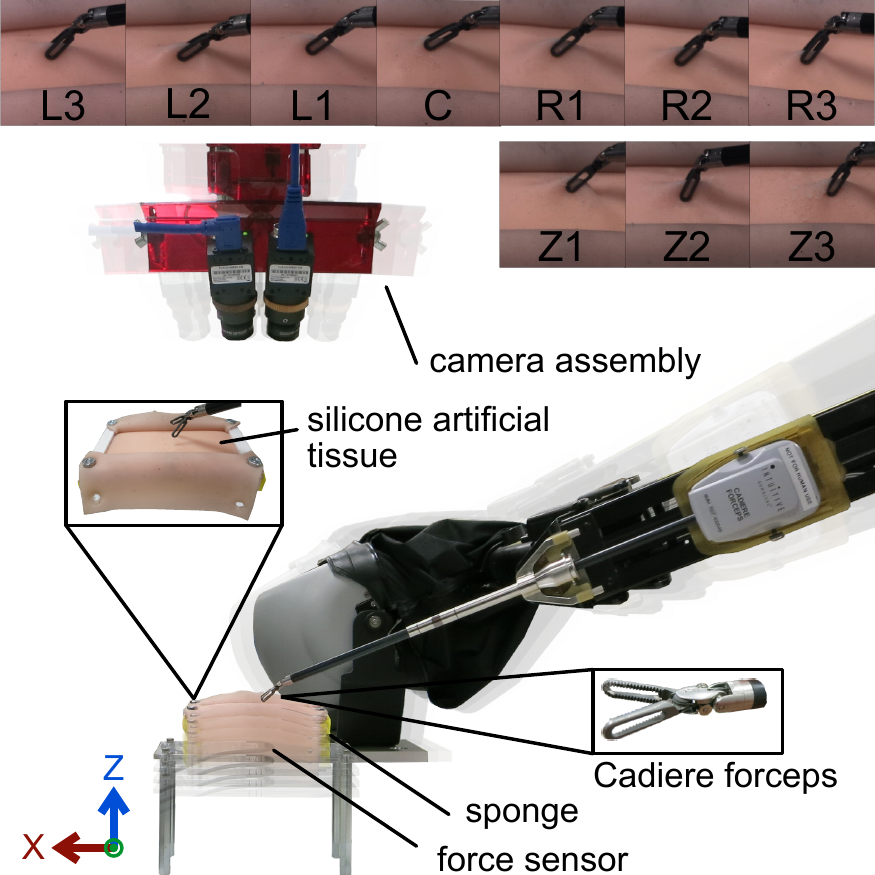}
\caption{Experimental set up comprising a da Vinci Research Kit equipped with Cadiere forceps, and silicone artificial tissue. The center (C) configuration is shown, with the other configurations, in which the camera assembly and robot based are simultaneously translated to the left (L) and right (R) relative to C by increasing distances (1-3), \updatedtext{and Z, where the stage is translated downwards by increasing distances (1-3), translucently overlaid. Z configuration positions are not shown to scale.} The image sequence displayed at the top shows the cropped camera image corresponding to each configuration.}
\label{experimentalsetup1}
\end{figure}

Here we introduce a monocular image and robot state-based, single time frame convolutional neural network for force estimation in RMIS. We investigated the accuracy and the generalizability of the full vision+state\,(V+S) network and its variants, vision-only\,(V) and state-only\,(S), to unseen viewpoints, joint configurations, tools, and materials, using a self-collected dataset consisting of palpations and pulls on silicone artificial tissue performed by a user teleoperating a da Vinci Research Kit \cite{kazanzides2014dvrk}. \updatedtext{To our knowledge, unlike prior works that only use position estimates \cite{aviles2016towards}\cite{Marban2019rnn}\cite{Jung2021}, ours is the first to use both kinematics (position and velocity measures) and force-torque state estimates as inputs to supplement vision inputs.} This allows us to study the relative importance of each class of features to the overall performance of the network. For comparison, we also present the results of a physics-based model \cite{Wang2019dynamic}, and a recurrent neural network (RNN) adapted from \cite{Marban2019rnn} as baselines.

We hypothesize that the V network will be sensitive to changes in viewpoint, but not to tool changes, while the S network will be sensitive to changes in joint configurations and tool, but not to changes in material. By combining both state and visual information, we expect the V+S network to match the performance of the best performing type of network over all unseen conditions. While recurrent networks can encode temporal information that should capture the effects of viscoelastic tissue properties on manipulation we reason that by incorporating the velocity estimates into a single time frame input, we will likely encode part of this information. We also reason that surgical movements usually occur on slow timescales that mitigate the contribution of viscoelasticity, such  that they are negligible with respect to network prediction error. These factors lead us to predict that the accuracy of our network will be comparable to that of the recurrent network, but with faster computation times, due to its reliance on single time frame inputs, that make it more suited for real-time use cases.


\section{METHODS}

\subsection{Hardware and Data Acquisition Setup}

A 125\,mm\,$\times$\,72\,mm piece of 3\,mm thick silicone artificial tissue (Limbs \& Things, Savannah, GA, USA) was placed on a 10\,mm thick sponge base and clamped to a 3D-printed platform. To create a consistent tissue interface border, the same type of silicone was pinned to the top and bottom of the clamping flange (Fig.\,\ref{experimentalsetup1} inset). This assembly was mounted to a six-axis Nano17 force-torque sensor (ATI Industrial Automation, Apex, NC, USA) to measure ground truth interaction forces and torques at 1\,kHz (Fig.\,\ref{experimentalsetup1}). Only the left camera in a stereoscopic camera assembly, consisting of two Flea3 cameras (FLIR Systems, Wilsonville, OR, USA), was used to capture 960\,$\times$\,540 pixel images at 30\,Hz. The camera assembly was mounted on a rail such that it could translate horizontally and pivot about its center. 


A da Vinci Research Kit (dVRK) with only the right hand controller and its corresponding patient side manipulator was used to perform the tissue manipulations. The default tool used was the Cadiere forceps (Fig.\,\ref{experimentalsetup1}). Both images from the stereoscopic camera assembly were shown to a human teleoperator through the stereoscopic display of the surgeon console. Similar to the camera assembly, the patient side manipulator was mounted on a rail that allowed for horizontal translations. Robot state information from the dVRK was also recorded at 200\,Hz and comprised of kinematic features as follows:
\begin{equation}
\centering
\begin{gathered}
    s_{\text{kin}} = [p_i,\,o_i,\,o_w,\,v_i,\,w_i,\,q_k,\,\dot{q}_k,\,q_{k,\text{des}}] \\
    i \in x,\,y,\,z \quad\quad k \in [1,7]
\end{gathered}
\label{pos_eqn}
\end{equation}
where $p$ is the world Cartesian position, $o$ the world orientation as a quarternion, $v$ the Cartesian velocity and $w$ the angular velocity in end-effector frame, $q$ the joint position, $\dot{q}$ the joint velocity, and $q_{\text{des}}$ the desired joint position. The robot state also included the following force-torque features:
\begin{equation}
\centering
\begin{gathered}
    s_{\text{force}} = [f_i,\,t_i,\,\tau_k,\,\tau_{k,\text{des}}] \\
    i \in x,\,y,\,z \quad\quad k \in [1,7]
\end{gathered}
\label{force_eqn}
\end{equation}
where $f$ is the Cartesian force in end-effector frame, $t$ is the torque in the end-effector frame, $\tau$ the joint torque, and $\tau_{\text{des}}$ the desired joint torque. This resulted in a full recorded state input vector in $\mathbb{R}^{54}$.

\begin{figure}[!t]
\vspace{1.5mm}
\centering
\includegraphics[width=0.65\linewidth]{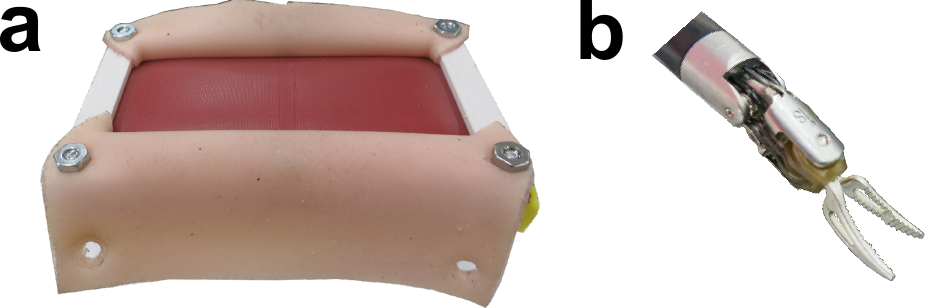}
\caption{Unseen environments used in testing. (a) The unseen material made of DragonSkin 10, with sponge underneath, mounted on the force sensor. (b) The unseen tool was the Maryland Bipolar forceps.}
\label{experimentalsetup2}
\end{figure}

To study our method's generalizability to an unseen material, the default silicone layer was replaced with a stiffer silicone layer made of Dragonskin 10 (Smooth-On Inc., Macungie, PA, USA) dyed red  (Fig.\,\ref{experimentalsetup2}a). The unseen tool for the experiments was the Maryland Bipolar forceps (Fig.\,\ref{experimentalsetup2}b).

\subsection{Data Collection}

\subsubsection{Experimental Conditions}
Using the dVRK equipped with the Cadiere forceps, a human teleoperator pulled and palpated the default artificial tissue to create two-minute video clips from different camera and robot base position configurations. There were a total of seven \updatedtext{main} configurations: the center (C) reference configuration, where the camera assembly was in line with the center of the artificial tissue, and the robot base was positioned 14.5\,cm to the right of the center of the tissue (negative X direction), and six configurations that comprised paired horizontal shifts of both the robot base and the camera assembly to the left (L), and the right (R) of the C configuration, by 1.5\,cm, 3\,cm and 4\,cm, corresponding to the 1, 2, and 3 positions as denoted in Fig. \ref{experimentalsetup1}. In the L and R configurations, the left camera of the camera assembly was oriented such that its field of view was centered on the middle of the artificial tissue sample. Additionally in the C configuration, data was collected for the unseen material using the Cadiere forceps, and also for the unseen tool using the default artificial tissue. For every video clip, the experimental set up was reset by moving the cameras and the robot base, dismounting the artificial tissue and swapping it out for another piece. \updatedtext{To help reduce overfitting to the workspace Z position, additional data was collected by lowering the height of the platform in quarter inch steps while in the C position to the Z1, Z2, and Z3 positions, corresponding with translations in the Z direction of -0.635\,cm, -1.27\,cm and -1.905\,cm, respectively (Fig.\,\ref{experimentalsetup1})}.


\subsubsection{Dataset Split and Preprocessing}
The training data was comprised of images, state inputs, and force labels, downsampled to 30\,Hz, from four video clips in each of the C, R2, and L2 configurations, \updatedtext{and two clips of Z2. During training, networks were validated on data from two clips in each of the C, R2, and L2 configurations, and a single clip in Z2. Two clips from each of the seen configurations C, R2, and L2, and a single clip from Z2 were included in the test dataset. Additionally, to test network generalizability, two clips from each of the R1, R3, L1, L3, unseen material, and unseen tool conditions, and a single clip each from Z1 and Z3 were used. This resulted in 49174 (33\%), 24659 (16\%), 24579 (16\%), 49241 (33\%) examples for training, validation, testing on seen, and unseen configurations, respectively.} To compare network and model performance across the same dataset, the first 80 and last 10 examples were removed from each test clip. This accommodated both the filtering needed for the physics-based model and the temporal history needed for the RNN. 


\renewcommand{\arraystretch}{1.25}

\begin{table*}
\vspace{1.5mm}
\centering
\caption{Average Root Mean Square Force Errors\,(N) Over all Axes for Test Conditions}
\label{rmse_table}
\begin{tabular*}{0.975\textwidth}{lS[table-format=1.3]S[table-format=1.3]S[table-format=1.3]S[table-format=1.3]S[table-format=1.3]S[table-format=1.3]S[table-format=1.3]S[table-format=1.3]S[table-format=1.3]S[table-format=1.3]S[table-format=1.3]S[table-format=1.3]S[table-format=1.3]} 
\toprule
                 & \multicolumn{11}{c}{\updatedtext{Workspace and Viewpoint} Configuration}                                                                                                                                                                                                                 & \multicolumn{1}{c}{\multirow{2}{*}{\begin{tabular}[c]{@{}c@{}}Unseen \\Material \end{tabular}}} & \multicolumn{1}{c}{\multirow{2}{*}{\begin{tabular}[c]{@{}c@{}}Unseen\\Tool \end{tabular}}}  \\ 
\cline{2-12}
                 & \multicolumn{1}{c}{L3} & \multicolumn{1}{c}{L2} & \multicolumn{1}{c}{L1} & \multicolumn{1}{c}{C} & \multicolumn{1}{c}{R1} & \multicolumn{1}{c}{R2} & \multicolumn{1}{c}{R3} & \multicolumn{1}{c}{Z1} & \multicolumn{1}{c}{Z2} & \multicolumn{1}{c}{Z3} & \multicolumn{1}{c}{Mean} & \multicolumn{1}{c}{}                                                                            & \multicolumn{1}{c}{}                                                                        \\ 
\hline
State-only                            & 0.317          & \textbf{0.288} & 0.299          & \textbf{0.306} & 0.247          & \textbf{0.244}  & \textbf{0.273} & 0.395          & 0.375          & \textbf{0.462}  & 0.321          & \hspace{0.5em}\textbf{0.521} & 0.517  \\
Vision-only                           & 0.521          & 0.499          & 0.504          & 0.496          & 0.456          & 0.499           & 0.465          & 0.721          & 0.728          & 0.714           & 0.560          & 1.837                        & 0.740  \\
Vision+State                          & \textbf{0.298} & 0.306          & \textbf{0.245} & 0.317          & \textbf{0.227} & 0.251           & 0.288          & \textbf{0.360} & \textbf{0.292} & 0.515           &\textbf{0.310}  & 1.067                        & \hspace{0.4em}\textbf{0.488}   \\
Vision+State RNN \cite{Marban2019rnn} & 0.451          & 0.411          & 0.401          & 0.453          & 0.354          & 0.367           & 0.386          & 0.583          & 0.594          & 0.632           & 0.464          & 0.900                        & 0.567    \\
Physics-based \cite{Wang2019dynamic}  & 0.957          & 0.786          & 0.748          & 0.713          & 0.741          & 0.655           & 0.720          & 0.862          & 0.747          & 0.927           & 0.786          & 0.851                        & 0.850   \\
\bottomrule
\end{tabular*}
\label{rmse_table}
\vspace{-1.5mm}
\end{table*}

All images were center-cropped to 300\,$\times$\,300 pixels and downscaled to 224\,$\times$\,224 pixels for input into the network. The images were also normalized using the mean and standard deviation of the ImageNet dataset. For the RNN benchmark implementation, the input images were preprocessed using the space-time image transformation with a spacing of 10 time steps, and normalization using the mean image of their corresponding video clip \cite{Marban2019rnn}.  

All state input features were normalized by element-wise subtraction of the mean and division by the standard deviation of the training dataset. While force and torque were recorded from the force sensor, only force was used as a ground truth for network training, as the torque measured at a location under the artificial tissue was not directly related to the torques at the end-effector, and provides limited value for feedback. The measured force was normalized in the same manner as the input features over the training dataset. 

\subsection{Network Architectures}

The V network was a ResNet50 \cite{he2016resnet} architecture pretrained on the ImageNet dataset. The last fully connected layer was modified to output a force estimate in 3 dimensions. Only the residual layers and fully connected layers were fine-tuned during training.

The image processing stage of the V+S network used the same ResNet50 architecture as the V network with one difference in that the fully connected layer had a 30-dimensional output. This output was concatenated with the 54-dimensional robot state and passed through 3 additional fully connected layers of size 84, 180, and 50, with a final 3-dimensional output. Batch normalization \cite{ioffe2015batch} and the ReLU activation \cite{krizhevsky2017imagenet} was used between these layers.

The S network comprised 6 fully connected layers of size 500, 1000\,(x3), 500, and 50, with a final 3-dimensional output. Batch normalization and the ReLU activation was used between layers. 

All networks were trained with mean squared error loss and L1 regularization for 100 epochs using Adam \cite{KingmaAdam}. After each training epoch, network accuracy was evaluated over the entire validation set. The set of network parameters that resulted in the highest accuracy on the validation set were used in the final trained network. A hyperparameter search over 20 epochs yielded a learning rate of 0.001 for the V+S and S network, and a learning rate of 0.0001 for the V network. All networks used a L1 regularization weight of 0.001. The size of the layers in our network were chosen empirically.

The benchmark RNN architecture \cite{Marban2019rnn} was adapted for comparison to the V+S network by swapping the VGG module with the same ResNet50 architecture used in the V network trained with mean squared error loss. The 2048-dimensional encoding from the ResNet was passed through the hyperbolic tangent activation before concatenation with a state input that was 54-dimensional, as opposed to 4-dimensional (Cartesian end-effector position and gripper state). The temporal sequence length was 60, and the training batch size was 250, with all other hyperparameters being the same as described in \cite{Marban2019rnn}.

\subsection{Metrics and Evaluation}

The accuracy of the regression output was measured using the root mean squared error (RMSE). This was calculated for each axis of force, with the mean over all axes also computed to provide an aggregate measure.

To evaluate the relative importance of the \updatedtext{kinematic} and force state inputs, feature removal studies were performed. Networks trained with \updatedtext{kinematic} features removed had the features described in Eqn. \ref{pos_eqn} set to 0, while those with force features removed had $q_{k,\text{des}}$ and features described in Eqn. \ref{force_eqn} set to 0. These networks were trained over 100 epochs with the same hyperparameters used to train the full models.

The comparison of real-time computational speed was done by simulating data preprocessing and inference through the V+S network and the RNN. For the RNN, the mean frame was computed online using exponential smoothing, and subtracted from the current frame  \cite{Marban2019rnn}. The space-time image was then created using previously stored and preprocessed frames. For both networks, image cropping, resizing, and normalization, and robot state input normalization were performed online. The simulation was performed one thousand times and and averaged to get the mean loop rate. The test was done in TensorFlow \cite{abadi2016tensorflow} with two Intel Xeon 2.2\,Ghz CPUs, and a NVIDIA Tesla T4 GPU. 

\section{RESULTS}

\begin{figure}[b!]
\centering
\includegraphics[width=0.9\linewidth]{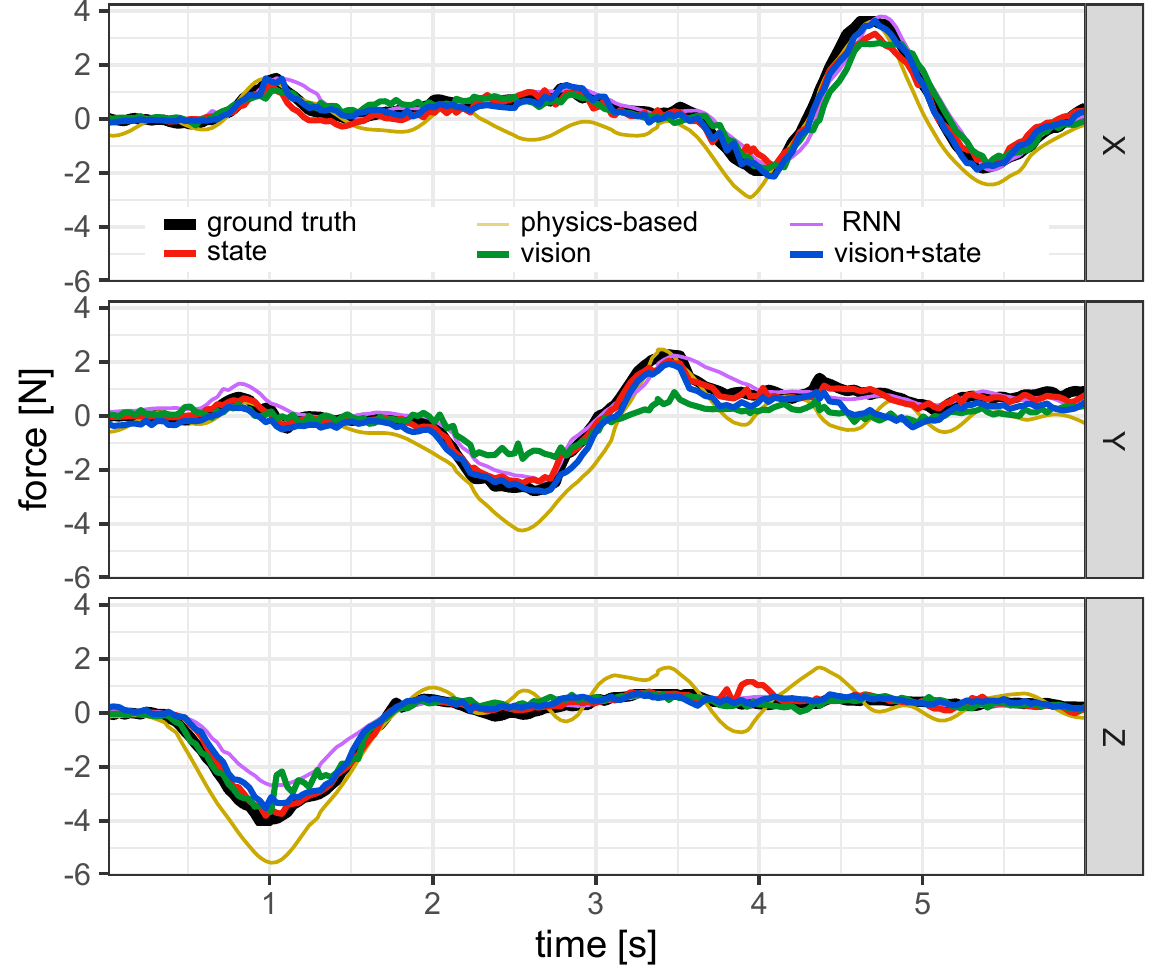}
\caption{Six-second excerpt of the predicted forces from each network for a center (C) configuration test trajectory in each axis of force.}
\label{traj}
\end{figure}

Table \ref{rmse_table} shows the mean RMSE of force in Newtons over all axes for the test set, for each test condition, averaged over the number of video sequences per condition. \updatedtext{The average RMSE over configurations indicate that for L3, L1, R1, Z1, and Z2, the V+S network achieved the lowest errors of 0.298, 0.245, 0.227, 0.360, and 0.292\,N, respectively. On L2, C, R2, R3, and Z3 , the S network achieved the lowest errors of 0.288, 0.306, 0.244, 0.273 and 0.462\,N, respectively. The V+S network had the best mean performance over all configurations. For the unseen material, the S network achieved the lowest RMSE (0.521\,N), while the V network had the highest error of (1.837\,N) due to its failure to predict variation in force, instead only predicting a bias. For the unseen tool, the V+S network performed best with a RMSE of 0.488\,N.} The physics-based model was consistently the least accurate in all conditions \updatedtext{except for the unseen material condition, where the networks with vision inputs had higher error.} Sample predictions of a force trajectory from the center (C) configuration are shown in Fig.\,\ref{traj}.


\begin{figure}[t]
\vspace{1.5mm}
\centering
\includegraphics[width=0.85\linewidth]{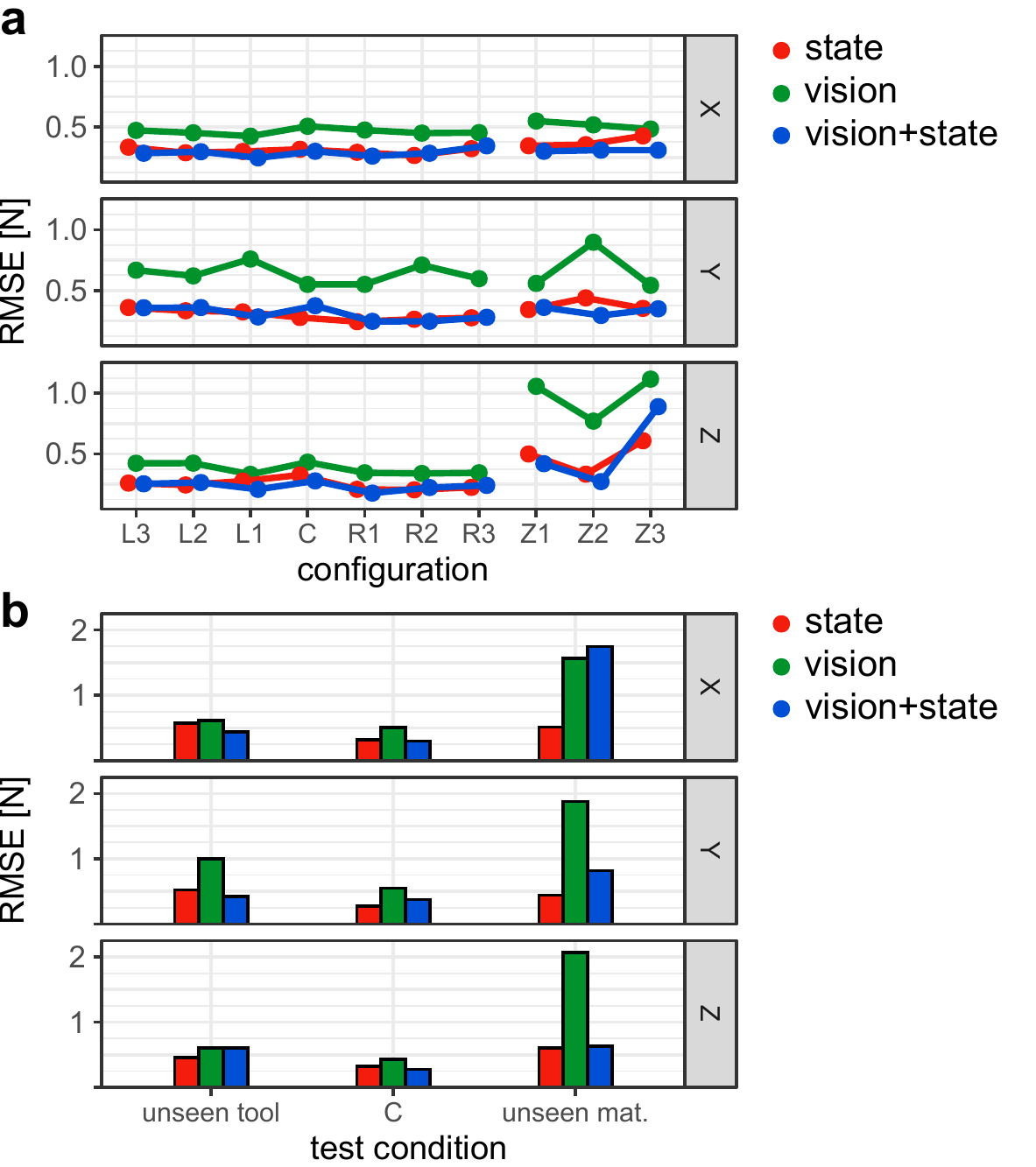}
\caption{(a) Comparison of the root mean square errors for each axis across tested configurations. (b) Comparison of the root mean square errors for each axis of the reference center condition (C), against the condition with the unseen tool, and the unseen material.}
\label{main_metrics}
\end{figure}

From a per-axis perspective, the V network achieved consistently higher errors over all test conditions, \updatedtext{with an exception in the X direction for the unseen material, where the V+S network had higher error (Fig.\,\ref{main_metrics}a and b). Between the Z configurations and the L to R configurations, the RMSE for the Z-axes for the S, V, and V+S models were noticeably higher for the Z configurations. This was because of the higher palpation forces found in that set relative to the L to R configurations.} For the unseen tool, the main contributors to the lower mean RMSE of the V+S network relative to the S network over all axes were the lower errors in the X- and Y-axes (Fig.\,\ref{main_metrics}b).


In all test conditions, the removal of either \updatedtext{kinematic} or force features from the input of the \updatedtext{S network resulted in higher error compared to the full state input. For the S network, the removal of force features consistently resulted in higher errors compared to the removal of \updatedtext{kinematic} features except in the Z direction for the unseen tool (Fig.\,\ref{ablate}). This was not consistently observed for the V+S network.}



The comparison of the loop rate of the V+S network against the RNN showed that the V+S network was faster. It had a mean loop rate of 37.7\,Hz, compared to 12.4\,Hz for the recurrent network.


\section{DISCUSSION}

\begin{figure}[t]
\centering
\vspace{1.5mm}
\includegraphics[width=0.7\linewidth]{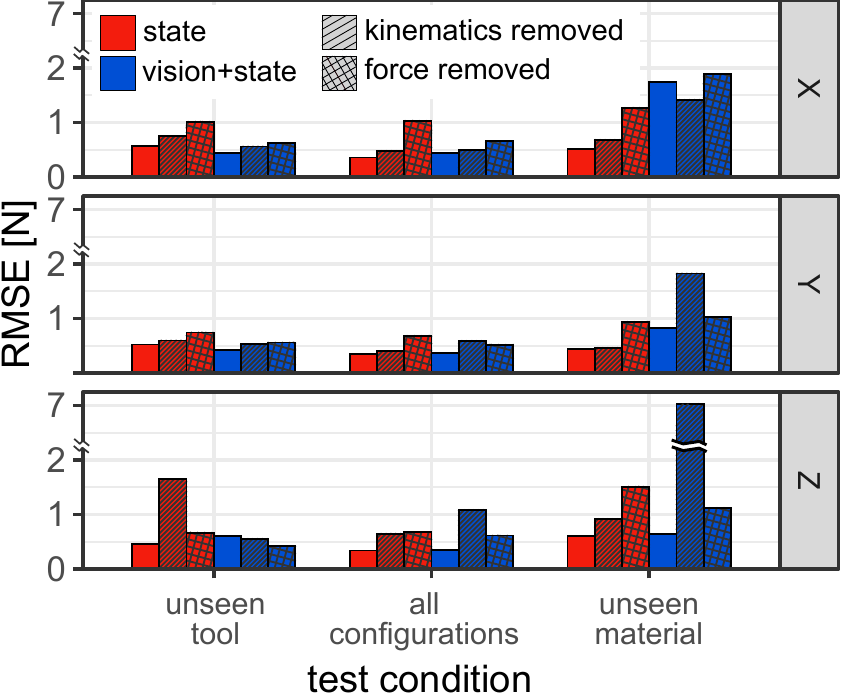}
\caption{Comparison of networks trained with removed features (\updatedtext{kinematic} or force features zeroed in the state input) against their original versions for different test conditions. The ``all configurations" condition is the average over all the camera and \updatedtext{workspace} configurations (L3 to Z3).}
\label{ablate}
\end{figure}

\updatedtext{The RMSE values in Table \ref{rmse_table} suggest that, compared to the V and S network, the V+S network had better average prediction accuracy across changes to configuration. However, for the Z configurations, the V+S network showed poorer performance in the Z3 configuration, which falls outside the range of Z-axis translation from C to Z2, compared to the S network. These observations suggest that, while the V+S network possibly overfits less to X-axis position of the workspace, there could be a tendency to overfit to Z-axis position. This is supported by the fact that across all configurations, the removal of \updatedtext{kinematic} features for the V+S network resulted in a higher RMSE in the Z-axis compared to removal of force features (Fig.\,\ref{ablate}). It is also likely that the visual inputs to the V+S network could have resulted in noise being introduced to the force estimates, where in some configurations, the V+S network tended towards the higher RMSE seen in the V network. This is supported by the observed increase of RMSE error of the V+S network for the unseen material, a scenario that causes the V network to fail (Fig.\,\ref{main_metrics}b).} 

\updatedtext{While all networks produced higher errors in the Z-axis for the Z configurations, large error in the V networks can be associated with the poor performance at high palpation forces due to reduced visual information at high compression. When excluding compression forces above 5\,N, the Z-axis RMSE for the V network for configurations Z1 to Z3 are within the range of 0.39 to 0.61\,N, which is close its range over L3 to R3. Meanwhile, the Z-axis RMSE for the V and V+S remain similar, implying lower sensitivity to Z-axis translation of the material than the V or V+S networks.} 

The failure of the V network for the unseen material highlights the adversarial effect of a new material, with unknown properties and unrecognized appearance, on vision-based networks. This effect also likely contributed to the poor performance of the RNN for the unseen material. For networks that only have state knowledge (S and physics-based), performance varied less relative to other conditions. 

\begin{figure}[t]
\centering
\vspace{1.5mm}
\includegraphics[width=0.8\linewidth]{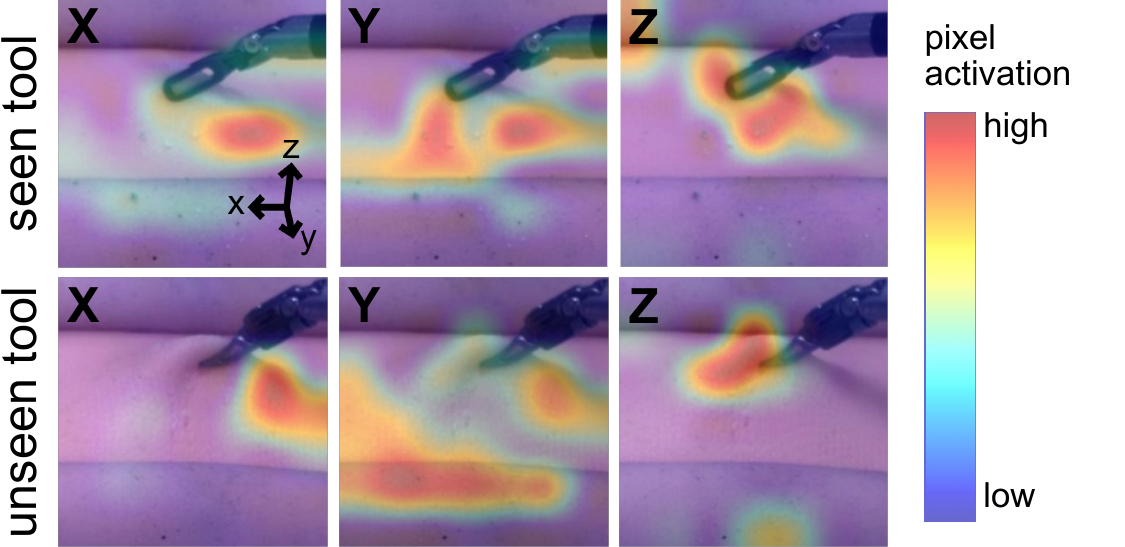}
\caption{Sample Grad-CAM visualizations from the V+S network in each axis for the seen tool (top row), and the unseen tool (bottom row).}
\label{heatmaps}
\end{figure}

One advantage of the V+S network is its better generalization to an unseen tool compared to the other networks. It is likely that the visual force cues from the environment help to make predictions less dependent on learning the tool dynamics, leading to improved performance. The different shape of the tool did not seem to confound the vision-based components of the networks, as the Grad-CAM heatmaps showed similar activations to those using the seen tool (Fig.\,\ref{heatmaps}). This was likely aided by the similar size and large contact areas of the forcep-type tools, which allow for material to be grasped in similar ways.


Together, the RMSE values confirm our hypothesis that the vision network shows sensitivity to changes in viewpoint, while partially refuting our hypothesis that the S network would show sensitivity to changes in robot configuration. \updatedtext{As shown in Fig.\,\ref{main_metrics}a, the sensitivity is pronounced in the Y-axis where the V network shows greater variability in RMSE compared to the S and V+S networks. This is consistent with the observation that the Y-axis is the direction in which there is the least depth perception in the image. This might have led to uncertainty in the visual force information in that direction, and under prediction of Y-axis forces as seen in Fig.\,\ref{traj}}. As predicted, the S network was more sensitive to an unseen tool relative to the V network, with the opposite trend for an unseen material. For the unseen tool, the V+S networks exceeded the performance of the partial networks. 

\updatedtext{As shown in Fig.\,\ref{main_metrics}, for all test conditions and axes, except for the unseen tool in the Z-axis for the V+S network, the full state results in the best accuracy. This suggests that interaction of the two types of information can be highly useful to force estimation deep learning networks. For example, if an inaccurate force estimate is available from the joint torques, the network could estimate and compensate for error using knowledge of the current kinematic state of the robot. For the exception in the Z-axis estimate for the unseen tool, the poorer performance of the full state network helps support the idea that a focus on visual deformation information helped the V+S network gain better accuracy when generalizing to a new tool. The feature removal studies for the networks with robot state inputs highlight the importance of force-based information for network accuracy. For the S network, removing force from the inputs resulted in higher errors in all axes, and in all conditions, compared to removing the \updatedtext{kinematic} inputs, with an exception being in the Z-axis of the unseen tool configuration (Fig.\,\ref{ablate}). This suggests that given only robot estimated force and torque inputs, the S network estimates Z-axis interaction forces in a tool specific manner. This trend of force input being more useful than \updatedtext{kinematic} inputs was not consistent for the V+S network, possibly owing to some position information being available visually to offset the removal of those features.}  


As predicted, the proposed V+S network showed comparable or better accuracy versus the RNN. While our experiments did not probe the reasons for this, it suggests that the increased state information compared to \cite{Marban2019rnn} could have potentially offset the need for temporal information. Furthermore, the influence of past information might result in a smoothing effect on predictions, \updatedtext{which could affect accuracy for fast variations in force. This is hinted at qualitatively from inspection of the force trajectories, for example in Fig.\,\ref{traj}, where at 3\,s the quick force variation in Y is not tracked well by the RNN.} However, it is important to note that the primary goal of the paper was to optimize the performance of our network, and it is likely that with more tuning and data, the RNN accuracy can be improved further. Because there is no need to serially process information like the RNN, nor perform temporal preprocessing of the input images, the static image V+S network exhibits faster real-time speeds than the RNN. With a loop rate above 30\,Hz, its speed is suitable for providing real-time visual feedback.

Because teleoperating the robot to collect data is time consuming, we did not collect a large enough dataset to evaluate statistical significance of the observed network differences. Thus, it is possible that there is undocumented variance in the performance of these networks. \updatedtext{More crucially, the small size of the training dataset could have also resulted in overfitting of the networks}, leading to its poorer generalization performance. In addition, there are several features of our work that constrain its applicability to actual RMIS scenarios. \updatedtext{We did not move the tissue manipulation platform in the Y-axis, and did so minimally in the Z-axis relative to the robot base and camera.} Thus the effect of such variations to network performance remains unanswered \updatedtext{and there remains concern that the S and V+S models overfit to workspace position.} In actual RMIS, the endoscope exhibits more movement, resulting in more varied viewpoints than used in our experiments. Most noticeably, the amount of zoom achievable by an endoscope is much greater. Additionally, the artificial tissues used in this study had relatively consistent appearances. Real biological tissues typically exhibit specularity and random visual blemishes that could either confound or enhance vision-based networks. \updatedtext{These concerns could potentially be addressed through the use of data set image augmentation techniques (e.g flipping, brightness and contrast adjustment) and training on varied zooms and viewpoints}. Lastly, our experiment presented the tissue in relation to consistent contextual markers that could be used by the network to estimate its viewpoint and configuration. In vivo, the contextual markers surrounding the manipulated tissue are more complicated and less distinct, presenting a potential challenge to network generalizability.

\section{CONCLUSION}

In this work, we presented a single time frame image- and robot state-based force estimation neural network, that while sensitive to changes in viewpoint when predicting forces in directions with reduced depth perception, can generalize to small changes in viewpoint and joint configuration, and unseen tools. \updatedtext{We also showed that \updatedtext{kinematic} information by itself is relatively less useful than force information estimated from motor currents, with the inclusion of both resulting in better network accuracy.} Using a larger input state, the single time frame network had similar accuracy to that of recurrent networks, with faster performance more suited for providing visual feedback.

While the networks are less accurate than distal force sensors, they do capture the larger trends in force variation. Thus, they are a promising direction in which to extend APMs with objective estimates of forces during surgical interactions, or to create sensing modules for autonomous application (e.g contact detection), in inanimate training \cite{Abbas2018} or potentially real, clinical environments. Future work will therefore investigate the suitability of our methods in providing objective feedback for training in inanimate environments, and also explore how the networks can be adapted for more demanding in vivo use, possibly through training on ex vivo or hydrogel-based tissue \cite{towner2019myomectomy}. 


\addtolength{\textheight}{-3.75cm}   



\section*{ACKNOWLEDGMENT}

The authors thank Negin Heravi for her ideas on CNN analysis, Diego Dall'Alba for providing the configuration files for the Maryland tool, and Arturo Marban for providing his image processing and TensorFlow RNN code. Compute was provided by the Stanford Research Computing Center.


\bibliographystyle{IEEEtran}
\bibliography{IEEEabrv,biblio}

\end{document}